# Deep Generative Model for Sparse Graphs using Text-Based Learning with Augmentation in Generative Examination Networks

Ruud van Deursen and Guillaume Godin

Firmenich SA, Research and Development, Rue des Jeunes 1, CH-1227 Les Acacias, Geneva, Switzerland, ruud.van.deursen@firmenich.com and guillaume.goding@formenich.com

**Abstract**

Graphs and networks are a key research tool for a variety of science fields, most notably chemistry, biology, engineering and social sciences. Modeling and generation of graphs with efficient sampling is a key challenge for graphs. In particular, the non-uniqueness, high dimensionality of the vertices and local dependencies of the edges may render the task challenging. We apply our recently introduced method, Generative Examination Networks (GENs) to create the first text-based generative graph models using one-line text formats as graph representation. In our GEN, a RNN-generative model for a one-line text format learns autonomously to predict the next available character. The training is stopped by an examination mechanism checking validating the percentage of valid graphs generated. We achieved moderate to high validity using dense g6 strings (random 67.8 +/- 0.6, canonical 99.1 +/- 0.2). Based on these results we have adapted the widely used SMILES representation for molecules to a new input format, which we call linear graph input (LGI). Apart from the benefits of a short compressible text-format, a major advantage include the possibility to randomize and augment the format. The generative models are evaluated for overall performance and for reconstruction of the property space. The results show that LGI strings are very well suited for machine-learning and that augmentation is essential for the performance of the model in terms of validity, uniqueness and novelty. Lastly, the format can address smaller and larger dataset of graphs and the format can be easily adapted to define another meaning of the characters used in the LGI-string and can address sparse graph problems in used in other fields of science.

**Introduction and related work.**

Graphs are used in a wide context of scientific fields, including chemistry, biology, engineering and social sciences. As a result from the widespread use of graphs, generative models of graphs are applied to many applications, such as modeling of physical and social interactions, generation of *de novo* molecules as well as constructing knowledge graphs. In the past, a variety of methods have been proposed for the development of generative graph models [1]. Based on these results, a key challenge for generative graph models has been identified: *learning* of generative models from a set of known graphs. This challenge plays a primary goal to improve the fidelity of generated graphs and it is a key application to generate new graph structures, e.g. new molecules, and to complete evolving graphs.

The earliest and most traditional generative graph models such as Barabasi-Albert, Kronecker graphs, exponential random graphs and stochastic block models [2-6] are frequently tailor-made to address a specific problem. As a consequence, the applicability domain of these models is restricted and the modeling strategy cannot be readily applied to other graph problems. Unfortunately, these models can frequently not learn from observed data, defining an essential shortcoming of the approach to address the challenge. An example of such a model is Barabasi-Albert model which has been carefully engineered to capture scale-free nature of empirical degree distributions, but which fails to capture many other aspects of real-world graphs such as community structure [4].

The recently introduced variational auto-encoders (VAE) [7] and generative adversarial networks (GAN) [8] were key contributions for the generative modeling of complex domains, such as learning of text and image data. Using these approaches, a variety of generative graph models have been introduced [9-12]. Simonovsky *et al.* have introduced a deep generative graph model based on VAE [11]. The major drawback of the method is the learning from a single graph [ref Kipf/Grover]. Another approach proposed by Li et al. [12], learned a deep generative model based on graph neural networks. The method is however restricted to relatively small graphs with 40 or fewer nodes. In summary, major challenges for generative graph models are thus the number of graphs and the size of the graphs that can be processed by the generative model. An additional key challenge for generative graph models includes the non-uniqueness of graphs. Recently, GraphRNN has been introduced as a first approach to address the challenges. The approach is based on learning the edges to the previously defined vertices and has been proven to outperform earlier models [13].

Indeed, as illustrated using graph enumeration, graphs are highly multi-dimensional and a combinatorial explosion is quickly observed [14]. Enumerative graph has been tailor-made to rapidly populate the space at the price of a combinatorial explosion. With the current hardware capabilities graphs larger than 19 vertices can hardly be assessed in a reasonable time or the task is reduced to a specialized task such as the enumeration of fullerenes [15]. A major disadvantage of exhaustive enumeration may be the strong dilution of the relevant space and discriminatory methods need to be applied to filter out all irrelevant graphs. Consequently, it has been proposed that is more efficient to generate graphs in the target space using a generative graph model [1].

In summary, the following major challenges have been defined. Firstly, learn to generate graphs using a (very) large existing set of graphs. Secondly, generate large and variable graphs. This point is especially challenging because a generative model for a graph with $n$ nodes must consider $n^2$ output values simultaneously. The number of $n$ nodes and $m$ edges may vary significantly between graphs and this needs to be addressed to efficiently generate relevant graphs for the space of interest. Thirdly, generating graphs is challenging because edge formation has complex dependencies. In real-word graphs, two nodes are more likely to be connected if they share common neighbors. Consequently, in a generative model one needs to address multiple edges simultaneously rather than a sequential series of independent events. Both the non-uniqueness and high complexity may require vast set of graphs to efficiently train a generative model. Lastly, non-unique representations are a key challenge for the general graph generation problem studied herein. In a deep generative model, we aim at creating a model that can autonomously learn the graph space and generate graphs with open-mindedness and vivid curiosity to efficiently explore the possible graph space. In other words, the generator should not be bound to a fixed set of constraints.

**Present work.**

The above challenges are not unique to generative graph modelling. In particular chemistry with the millions of known molecules, are open to the same challenges [16]. Previously, we have successfully introduced generative examination networks (GEN) which take the one-line text string SMILES as input for a RNN-based generative model to create new molecules. GENs combine a network of an autonomously learning generative neural network combined with an examination method to prevent the model from overfitting and memorizing the training dataset. In GENs, training is stopped early as soon as the generative model has deduced sufficient knowledge from the training set to reconstruct the property space with new valid text strings. Our previous results on SMILES clearly show that the generators can independently learn the used alphabet and grammar. Most previous work on discovering molecule structures make use of a expert-crafted sequence representations of molecular graph structures (SMILES) [17-19]. Most recently, SD-VAE [20] introduced a grammar-based approach to generate structured data, including molecules and parse trees. In contrast to these works, we consider the fully general graph generation setting without assuming features or special structures of graphs. The obtained generators stand out by a high degree of novelty while keeping a focus on the property space of the training set. If needed, smaller datasets can be easily augmented with randomized SMILES to obtain a better train generator.

Graphs are frequently stored using the dense one line text format g6 (GRAPH6), which is a transformation of the adjacency to a linear sequence of characters [21,22]. Primary benefit of g6-strings is the strict forward encoding of the characters, defining an ideal input system for LSTM-type recurrent neural networks [23,24]. A known disadvantage of encoding the adjacency matrix is the large number of permutations possible. Consequently, we hypothesize that g6-strings are not readily generalizable for machine-learning and will be outperformed by the dataset with the canonical g6-strings. This effect is likely to increase with increasing sparsity of the graph. Herein we also introduce an alternative one line text format to represent unweighted undirected sparse graphs. This format is based on SMILES string and we have changed the atom symbols to a graph symbol defining the degree of the vertex. We call this new graph representation linear graph input (LGI). In LGI-format vertices are printed in a depth-first tree traversal of the graph, intrinsically conserving the graph topology in the text representation. Furthermore, data augmentation is already available in SMILES and can thus be applied for the LGI-format. Augmentation has been shown to be beneficial for text learning in chemistry [25-30].

Here we introduce g6-GEN and LGI-GEN as two methods to create a generative model for the generation of graphs using g6 or LGI strings. In those two GENs, recurrent neural network used for the generator used by self-directed learning, while an online examination mechanism periodically checks the learning progress of the generator. Here we use online statistical quality control (SQC) [31] to assess the knowledge of the network based on the percentage of valid generated graphs. To our knowledge g6-GEN and LGI-GEN are the first ever attempts to use a text-based approach for a generative model of graphs. The approach is also open to be trained with (very) large sets of existing graphs.

We observed that g6 is not well adapted for data augmentation. We believe that these results are an immediate consequence of the fact that g6 string encode the adjacency matrices and the number of matrix permutations is sheer endless, yielding a vast set of isomorphically identical text strings. In particular for sparse graphs there is a combinatorial explosion of possibilities to assign the edges to the matrix. Initial results show using g6-strings, show modest to high validity using randomized strings (67.8 +/- 0.6) and canonical (99.1 +/- 0.2) (see figure SI1). These results are in line with the hypothesis that encoding the adjacency matrix is not efficient as a result of the high degree of permutability of the adjacency matrix. Based on these results on the g6-string, research was continued with our newly introduced LGI-format.

To demonstrate the generative performance of our model, we have used a large chemical dataset extracted from PubChem [ref PubChem]. The dataset contains 106k unique graphs which are realistic fragments typically used in pharmaceutical and olfactive industries. The calculations show that LGI-GEN can adequately address the aforementioned challenges and that augmentation is essential to improve the models.

**Methods**

**LGI-format.** Albeit g6-strings define an easy-accessible strictly linear format, we herein evaluated the use of the aforementioned "linear graph input" (LGI) format (Table 1). The syntax was derived from the molecular Simplified Molecular Input Line Entry System (SMILES) system [32], which write the nodes using a depth-first tree traversal of the molecular graph. The LGI-format is a format centralized around the vertices. In this format we used the ASCII characters {@-F} to encode the degrees {0-6} observed for the vertices. We used numerical values to indicate neighboring cyclic vertices. These values were written immediately after the degree character of the vertex and always come in pairs to indicate the neighboring vertices. Ring indices composed of double digits were preceded by the character "%". If a graph is part of two cycles simultaneously, the vertex may be followed by multiple ring indices, e.g. C12 defines a 3-degree vertex in rings 1 and 2 of the graph. In case of branches, the characters "(" and ")" were used for branch opening and closing, respectively. These brackets always come in pairs.

**Table 1: Conventions used in the LGI-format.** The format is derived from the one-line text format SMILES widely used in chemistry [32].

| Characters[1] | Described property | Note |
|---|---|---|
| @, A, B, C, D, E, F | Degrees 0, 1, 2, 3, 4, 5, 6 | One character/degree |
| 1, 2, 3, 4, 5, 6, 7, 8, 9 | Ring indices | In pairs; Reusable[2] |
| ( , ) | Branch opening and closing | In pairs; Reusable[3] |
| % | Ring indices character for digits | Occurs with >9 rings simultaneously |

1) Selection of characters for the description are open to any kind selected by the user.
2) Ring indices are defines as pairs and can be reused (see example in figure 1E).
3) Branch opening/closing characters come in pairs "(" and ")".

**Graph randomization and augmentation.** For g6-string, string is converted into a networkx Graph, graph is randomized by shuffling the order of the vertices and written into a new g6-string encoding the new adjacency matrix. The LGI-strings were converted into the SMILES equivalent, randomized using SMILES in RDKit with *dorandom=True* and written into a new LGI-string [33]. Augmentation was performed by repeating the randomization attempts. The generated strings were subsequently deduplicated to create a unique set of strings for each graph (Figure 1).

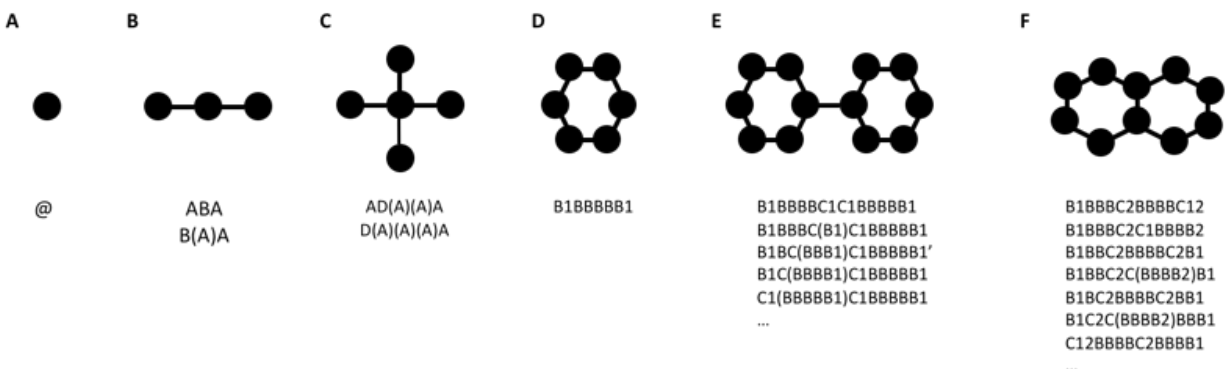

**Figure 1: LGI-format for a set of example graphs.** The graphs are displayed with multiple lines to indicate an alternate random LGI representing the identical graph. Graphs A and D generated only 1 possible LGI-string. Graph B and C generated 2 LGI-strings. For graphs E and F a subset of possible LGI-strings is shown.

**Dataset preparation for molecular datasets.** Pubchem [34] was downloaded on March 2019. The canonical SMILES string *PUBCHEM_OPENEYE_CAN_SMILES* was extracted, split into fragments and converted into canonical SMILES using RDKit version 2019.03.3 [33,35]. Only organic molecules, i.e. those that contain at least one carbon and all other atoms are part of the subset H, B, C, N, O, F, S, Cl, Br or I, were retained and then deduplicated to produce a set of unique SMILES. From this dataset, we extracted a representative set of 225k fragment-sized molecules typically explored in the pharmaceutical and olfactive industries [36,37]. The molecules were subsequently converted to an undirected unweighted graph *G(V,E)* and written as g6- or LGI-string. The LGI-strings were randomized by shuffling the vertex in a random order. Augmentation of the LGI-strings was performed using repeated randomizations. All datasets were deduplicated to create a set of unique LGI-strings. The final training set contained 103,826 graphs, 15,813 unique scaffolds and 1,321 unique ring systems.

**Architecture.** Modeling has been performed using neural networks based on python libraries tensorflow [38] and Keras [39]. The method has been coded using the Python [40]. All evaluated deep generative models were composed of a single LSTM-layer of size 128 followed by four parallel LSTM-layers of size 64 (figure 2A) [27]. These parts of the network we call embedding and encoding layers, respectively. The parallel encoding layers were normalized using LayerNormalization and merged by concatenation. The output consists of a single Dense layer with a size equal to the number of possible ASCII characters. and two composed consecutive LSTM layers, which we call embedding and encoding layers, respectively. A single architecture with biLSTM layers was evaluated using an embedding layer of size 64 and encoding layers of size 32 to generate a network with the identical number of hyperparameters. The code is freely available under a clause-3 BSD license [41].

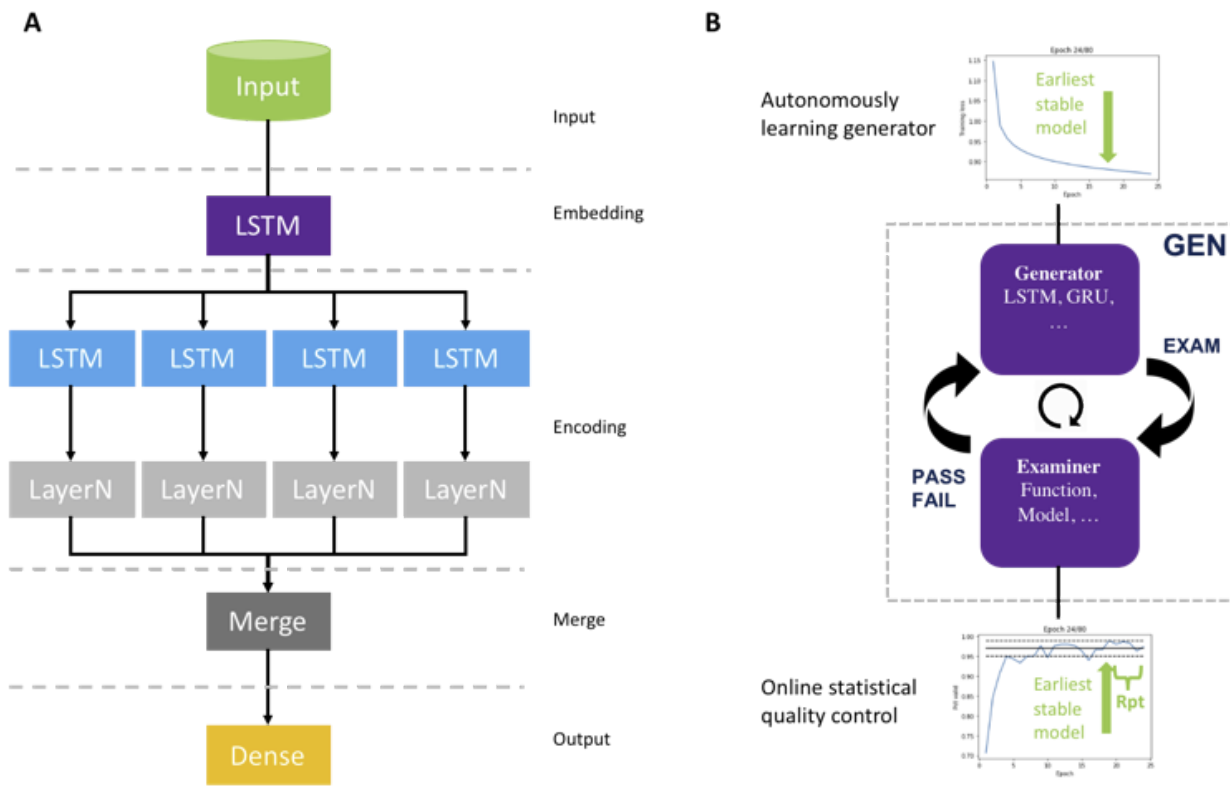

**Figure 2: Used architecture for the generator in the generative examination network [27].** A) Evaluated architecture for the RNN model, composed of an input followed by two consecutive LSTM-layers, which we call embedding and encoding layers, respectively. The output of the encoding layers is merge by concatenation. B) Training framework of a generative examination network (GEN). The autonomously learning generator passes periodically an exam using a statistical sample of generated graphs. The training is stopped early after the generator passed a consecutive series of exams ("*Rpt*"). The earliest stable model is used for graph generation.

**Training of the generative models.** LSTM is characterized by a conservative long-range memory [23]. In order to improve the degree of generativity and overall performance of the model we have trained all models with a randomized g6-string or LGI-string or with multiple randomized g6-strings or LGI-string using graph augmentation as described above. A comparison was made to an architecture trained with a set of canonical g6- and LGI-strings. We embedded the generator in a generative examination network (GEN, figure 2B). In GENs, an autonomously learning neural network for the generative model is combined with an independent online examination mechanism. Here we trained the generative model using categorical cross-entropy to predict the next character. To avoid overfitting the training set, neural networks are typically subjected to an early stopping mechanism. Here we modified the existing Keras' Callback function EarlyStopping to become a method applying an online statistical quality control (SQC) on the percentage of valid generated graphs after every epoch. The callback function was parameterized with a target percentage of valid graphs, the number of a statistical sample (**Nsample**) and a waiting period (patience). Prior to training, the callback function computed the upper and lower margin for the target percentage and sample size using the 95% confidence interval [42]. After every epoch, the callback function asked the generative model to generate *Nsamples* g6-strings and subsequently measured the percentage of validity. The callback function stopped training early after the model showed stable generation results for 10 consecutive epochs (or as indicated by the user defining the value for *patience*). The counter for the early stopping function was reset whenever the percentage of valid graphs fell below the lower margin of the 95% confidence interval. Upon completion the early stopping function selected the earliest available model weights to keep the degree of selectivity.

**Graph validity.** For g6 strings, graph validity has been measured by testing for the correct length of the g6-string, i.e. 4 characters for a graph with 6 vertices (EhEG). Additionally, a graph was considered invalid if one or more vertices had a degree which was not observed in the training set and thus out-of-domain [43]. Graphs generated with LGI-strings were considered valid if the generated string had a matching set of ring indices and branch opening/closing characters as well as all vertices had a degree corresponding to the LGI-character (i.e. a vertex with character B had 2 edges). Validation of the LGI-strings was thus considerably easier because the expected degree is embedded in the format.

**Property evaluation.** The property space of the graphs in the training set and generated sets was evaluated using the length of the LGI-strings, number of nodes in the graph and graph energy (GE), summing the absolute values of the eigendecomposition of the adjacency matrix (equation 1):

**Equation 1:**

$$GE = \sum_i |\lambda_i| \quad (1)$$

For all properties the overlap between the distributions was measured using the continuous Tanimoto coefficient (equation 2) and Jensen - Shannon Divergence coefficient (JSD, equation 3):

**Equation 2:**

$$T(A,B) = \frac{\sum_i A_i B_i}{\sum_i A_i^2 + \sum_i B_i^2 - \sum_i A_i B_i} \times 100\%$$

**Equation 3:**

$$JSD(A,B) = H\left(\sum_{d \in \{A,B\}} a_i d_i\right) - \left(\sum_{d \in \{A,B\}} a_i H d_i\right)$$

**Evaluation of the generated graphs.** Globally, the set of generated graphs was evaluated using the percentage of validity, percentage of uniqueness and the percentage of novelty, applying equations 4, 5 and 6, respectively:

**Equation 4:**

$$Validity = \frac{\|valid\|}{\|generated\|} \cdot 100\%$$

**Equation 5:**

$$Uniqueness = \frac{\|unique\|}{\|valid\|} \cdot 100\%$$

**Equation 6:**

$$Novelty = \frac{\|unknown\|}{\|unique\|} \cdot 100\%$$

Where *valid* defines the number of valid graphs, *generated* the number of generated g6-strings, *unique* the number of unique graphs as checked by isomorphism and *unknown* the number of graphs unknown in the training set.

**Scaffold and ring systems.** All graphs in the training and generated sets were also evaluated on the level of scaffold and ring systems. For a graph *G(V,E)*, the scaffold was generated by identification of all cyclic vertices based on the minimal cycle basis (MBC). The graph keeping all cyclic vertices and connected acyclic neighbors. The ring system for every graph G(V,E) was obtained keeping all cyclic vertices. If a graph contained multiple connected subgraphs, these graphs were considered separate scaffolds and ring systems. Fully acyclic graphs were excluded from the analysis.

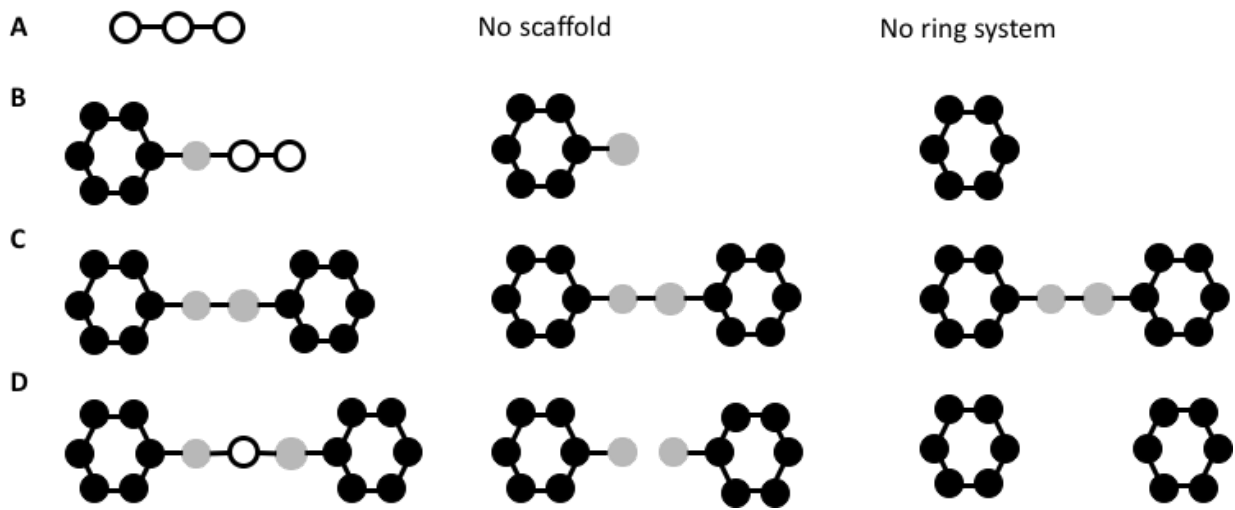

**Figure 3: Scaffold and ring system definition.** For scaffold analysis all cyclic vertices (solid black dot) and adjacent acyclic vertices were kept. For an analysis of ring systems only cyclic vertices were kept. All other acyclic vertices (white dot) were removed. A) Fully acyclic graphs have neither a scaffold nor a ring system and were not considered in the evaluation; D) Graph with multiple disconnected subgraphs were split into individual graphs after removal of acyclic vertices.

**Results**

In our first calculations we have evaluated the possibility to use g6-strings as input format (figure SI1). The results clearly show that the model can only learn if the g6-strings are in a canonical form (99.1 +/- 0.2% validity). The randomized form displayed a significantly lower performance (67.8 +/- 0.6% validity). These results encouraged us to introduce the lgi-format and all following results are exclusively based on the lgi-format.

A first evaluation of the model was performed evaluating the overall performance of the generated graphs by each generative model (table 2, figure 3). The results for the canonical form with the LSTM-LSTM or biLSTM-biLSTM was stable after 6 and 9+/-1 epochs, respectively. The percentage of validity, uniqueness and novelty furthermore suggested that the use of bidirectional LSTM layers does not yield an improvement on the overall performance. On the contrary, the performance for uniques (-1.0%) and novelty (-1.3%) even dropped slightly. Consequently, we have evaluated the effect of LGI-randomization and LGI-augmentation exclusively using the architecture with LSTM-layers.

**Table 2: Summary tables on the performance of the graph generator.** All architectures were independently trained three times.

| Model / Dataset | $N_{Embedding}$ | $N_{Encoding}$ | Epoch[1] | Valid | Unique | Novelty[2] |
|---|---|---|---|---|---|---|
| LSTM-LSTM Canonical | 128 | 64 | 6, 6, 6 | 97.2 +/- 0.1 | 94.0 +/- 0.1 | 64.6 +/- 0.4 |
| biLSTM-biLSTM Canonical | 64 | 32 | 9, 10, 8 | 97.2 +/- 0.1 | 93.0 +/- 0.1 | 63.3 +/- 1.5 |
| LSTM-LSTM Random 1x | 128 | 64 | 6, 6, 6 | 98.6 +/- 0.1 | 93.4 +/- 0.3 | 64.1 +/- 0.6 |
| LSTM-LSTM Random 2x | 128 | 64 | 6, 6, 6 | 98.9 +/- 0.1 | 87.8 +/- 0.4 | 53.7 +/- 0.3 |
| LSTM-LSTM Random 3x | 128 | 64 | 6, 6, 6 | 99.3 +/- 0.1 | 92.6 +/- 0.5 | 58.6 +/- 0.4 |
| LSTM-LSTM Random 4x | 128 | 64 | 6, 6, 6 | 99.3 +/- 0.1 | 92.2 +/- 0.1 | 58.8 +/- 0.4 |
| LSTM-LSTM Random 5x | 128 | 64 | 4, 4, 4 | 99.5 +/- 0.1 | 93.5 +/- 0.2 | 60.1 +/- 0.5 |
| LSTM-LSTM Random 10x | 128 | 64 | 3, 3, 3 | 99.4 +/- 0.1 | 95.5 +/- 0.2 | 63.6 +/- 0.3 |

1) Selected models for generation.
2) Novelty defines the generated graphs unknown in the training set.

The results for the overall performance show clearly that randomization and augmentation is useful. Firstly, at 5-fold and 10-fold augmentation the training could be stopped earlier. Secondly, the validity of all calculations is higher than the validity of the canonical form and reached 99.3% for all calculations performed with 3 or more augmentations of the dataset. Thirdly, we observed a performance drop of the percentages uniqueness and percentage novelty during the augmentation. Indeed, the lowest values are observed for a 2-fold augmentation with a drop of 6.6% and 10.4% for the percentages of uniqueness and novelty, respectively. With an increasing degree of augmentation, the percentages of uniqueness and novelty increase steadily to a level equal to the canonical form.

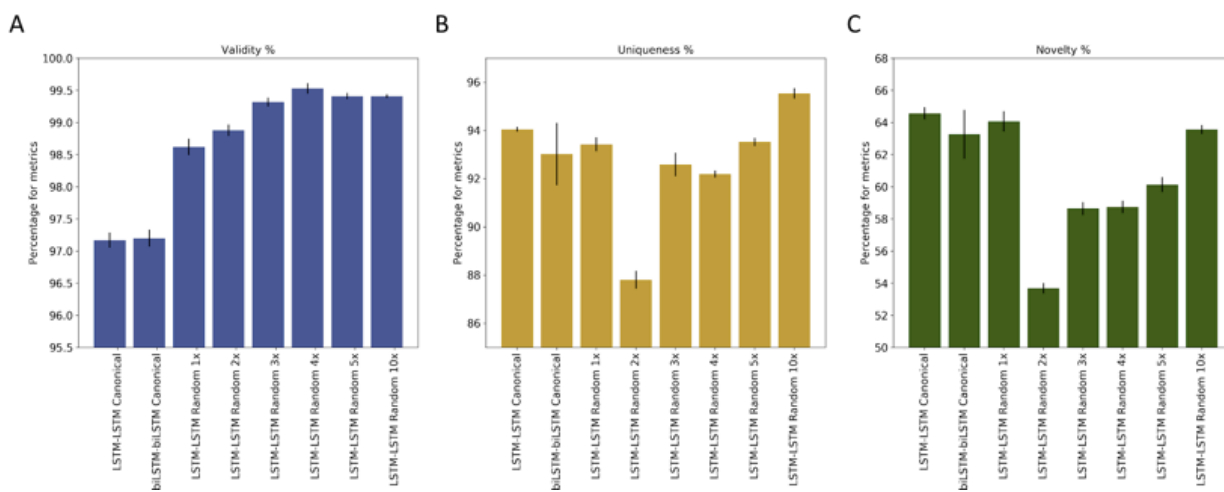

**Figure 4: Summary figure with generation performance.** See methods section for calculation details. Results shown for A) Percentage of valid generated graphs. B) Percentage of unique generated graphs. C) Percentage of novelty, i.e. new graphs not known to the dataset.

A more detailed analysis was performed evaluating the property distributions of the generated datasets (table 3, figure 5, figures SI2-SI4). The results, in particular the Jensen-Shannon Divergence (JSD) coefficient, show that randomization and augmentation significantly improves the output of the generative models. Indeed, at a level of 3 or more augmentations, the values for JSD are significantly lower (-0.043) than for generative models based on either the canonical form or a limited number of 1 or 2 augmentations. Using the Tanimoto coefficient for distribution overlap we observe that randomization improves all properties. The LGI-length gradually improves from 91% to 99%, the number of nodes $||V||$ in the graph improved from 85.6% to 98.8% and the graph energy GE improved from 93.6 tot

99.1%. The best results were observed for 3-10 fold augmentation, which appears to be a result of more natural input distributions for the text strings (figure SI2).

Table 3: Property comparison between generated graphs to the training set.

| Model / Dataset | LGI Length | | Number of nodes ‖V‖ | | Graph Energy GE | |
|---|---|---|---|---|---|---|
| | Tanimoto[1] | JSD[2] | Tanimoto[1] | JSD[2] | Tanimoto[1] | JSD[2] |
| LSTM-LSTM Canonical | 91.0 +/- 0.4 | 0.129 +/- 0.003 | 85.9 +/- 0.4 | 0.156 +/- 0.002 | 93.6 +/- 0.2 | 0.101 +/- 0.002 |
| biLSTM-biLSTM Canonical | 87.7 +/- 4.3 | 0.154 +/- 0.033 | 83.4 +/- 3.5 | 0.168 +/- 0.017 | 94.2 +/- 0.8 | 0.097 +/- 0.004 |
| LSTM-LSTM Random 1x | 97.4 +/- 0.3 | 0.080 +/- 0.004 | 96.9 +/- 0.2 | 0.142 +/- 0.004 | 97.4 +/- 0.2 | 0.093 +/- 0.003 |
| LSTM-LSTM Random 2x | 96.4 +/- 0.3 | 0.084 +/- 0.003 | 91.8 +/- 0.2 | 0.129 +/- 0.002 | 94.5 +/- 0.3 | 0.091 +/- 0.003 |
| LSTM-LSTM Random 3x | 99.4 +/- 0.1 | 0.045 +/- 0.002 | 99.9 +/- 0.1 | 0.092 +/- 0.003 | 99.0 +/- 0.2 | 0.048 +/- 0.005 |
| LSTM-LSTM Random 4x | 99.3 +/- 0.1 | 0.049 +/- 0.002 | 98.9 +/- 0.2 | 0.100 +/- 0.004 | 99.2 +/- 0.2 | 0.051 +/- 0.004 |
| LSTM-LSTM Random 5x | 98.9 +/- 0.3 | 0.049 +/- 0.003 | 97.4 +/- 0.2 | 0.089 +/- 0.001 | 98.6 +/- 0.3 | 0.052 +/- 0.002 |
| LSTM-LSTM Random 10x | 99.3 +/- 0.2 | 0.041 +/- 0.003 | 98.8 +/- 0.2 | 0.093 +/- 0.003 | 99.1 +/- 0.1 | 0.046 +/- 0.001 |

1) Continuous Tanimoto calculated using equation 2.
2) Jensen-Shannon divergence calculated using equation 3.

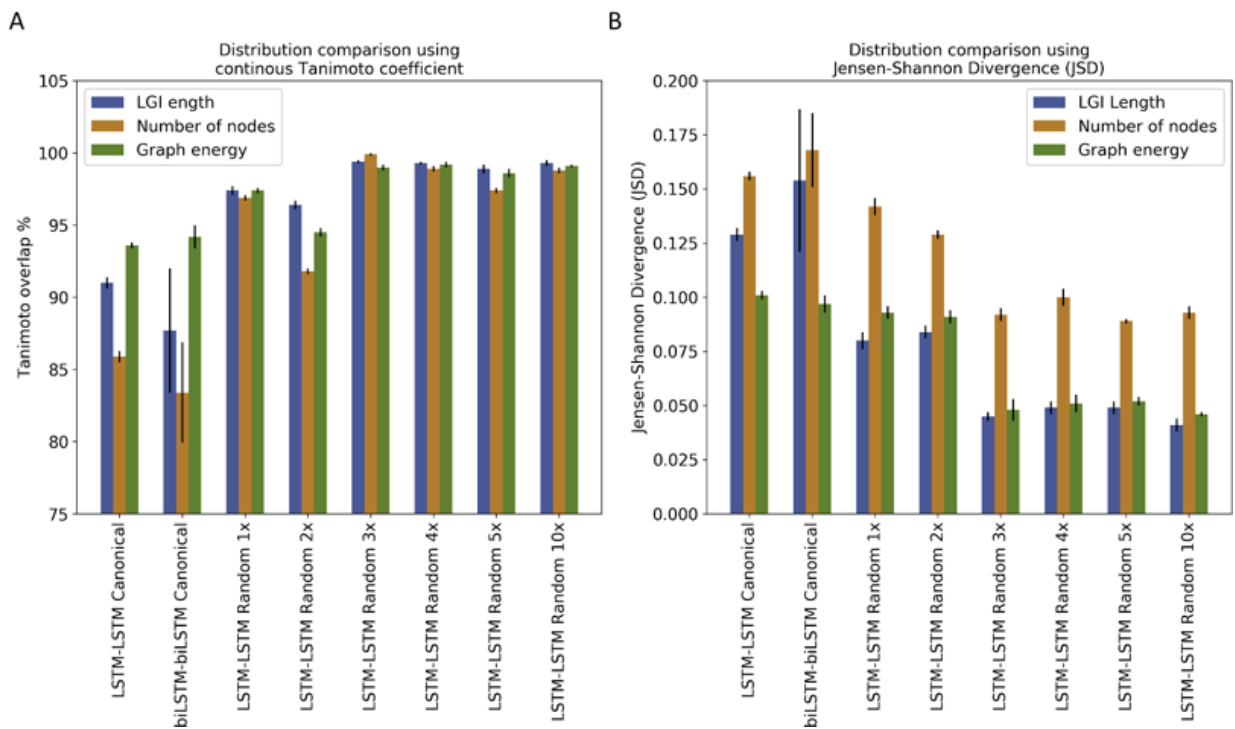

Figure 5: Summary of the property distributions. A) Results shown for the continuous Tanimoto similarity coefficient. B) Results shown for the Jensen-Shannon convergence.

Next, we have also performed an analysis on the generated scaffolds and rings in the generative models (see methods for definition). The results are summarized (Table 4 and Figures 6 and SI5) and show more evidence that augmentation is beneficial for graph generation. In the table we have computed the number of generated graphs needed to reach an equally large set of unique graphs, scaffold or ring systems. This point we call herein *intersection* with the training set. Firstly, we observe that the generative model needs a longer time to reach the same number of unique graphs and thus occasionally generates duplicates. The results also show that augmentation first increases this intersection point with a peak at 2-fold randomization, followed by a strong decreasing trend with increasing number of augmentations. These results are in line with the earlier observations that at least 3-fold augmentation is needed to get good generative results on the property distributions.

**Table 4: Intersection points of graphs, scaffold and ring systems to reach the size of the training set.** Training set contained 103,826 graphs, 15,813 unique scaffolds and 1,321 unique ring systems.

| Architecure | Augmentation | Graphs | | Scaffolds | | Ringsystems | |
|---|---|---|---|---|---|---|---|
| | | Unique[1] | New[1] | Unique[1] | New[1] | Unique[1] | New[1] |
| LSTM-LSTM | Canonical | 155,531 (1.00) | 213,871 (1.00) | 115,640 (1.00) | 183,302 (1.00) | 89,506 (1.00) | 195,163 (1.00) |
| biLSTM-biLSTM | Canonical | 174,945 (1.12) | 252,746 (1.18) | 118,955 (1.08) | 197,901 (1.08) | 95,096 (1.06) | 220,457 (1.13) |
| LSTM-LSTM | Random 1x | 147,824 (0.95) | 197,081 (0.92) | 95,735 (0.80) | 145,764 (0.80) | 66,045 (0.74) | 134,014 (0.69) |
| LSTM-LSTM | Random 2x | 179,762 (1.27) | 257,118 (1.20) | 104,800 (0.90) | 165,636 (0.90) | 75,924 (0.85) | 167,576 (0.86) |
| LSTM-LSTM | Random 3x | 168,767 (1.09) | 228,332 (1.02) | 117,903 (1.05) | 192,455 (1.05) | 100,107 (1.12) | 213,051 (1.09) |
| LSTM-LSTM | Random 4x | 169,240 (1.09) | 228,980 (1.03) | 119,396 (1.04) | 191,349 (1.04) | 96,397 (1.08) | 200,679 (1.03) |
| LSTM-LSTM | Random 5x | 158,890 (1.02) | 218,530 (1.02) | 89,617 (0.77) | 141,535 (0.77) | 63,465 (0.71) | 126,058 (0.65) |
| LSTM-LSTM | Random 10x | 149,099 (0.96) | 198,806 (0.93) | 87,695 (0.76) | 138,551 (0.76) | 56,321 (0.63) | 115,215 (0.59) |

1) Relative time indicated in brackets; measured against LSTM-LSTM with canonical graphs.

Secondly and more importantly, we observe significant improvements for the discovery of new scaffolds and rings with augmentation. We observe that the number of generated graphs needed is lower than the number needed using the canonical form (figure 6B-C). More strikingly, the intersection for the augmented datasets is also lower than the size of the training set. The best results were obtained for 10-fold augmentation, which created 15,813 unique scaffolds in a set of 87,695 scaffolds and 1,321 unique ring systems in a set of 56,321 generated graphs, respectively. These values are particularly remarkable, because the number of generated graphs needed for the discovery are significantly smaller than the size of the training set (103,826) used to train the model, suggesting the model is an excellent explorer for new scaffolds and ring systems (figure 7).

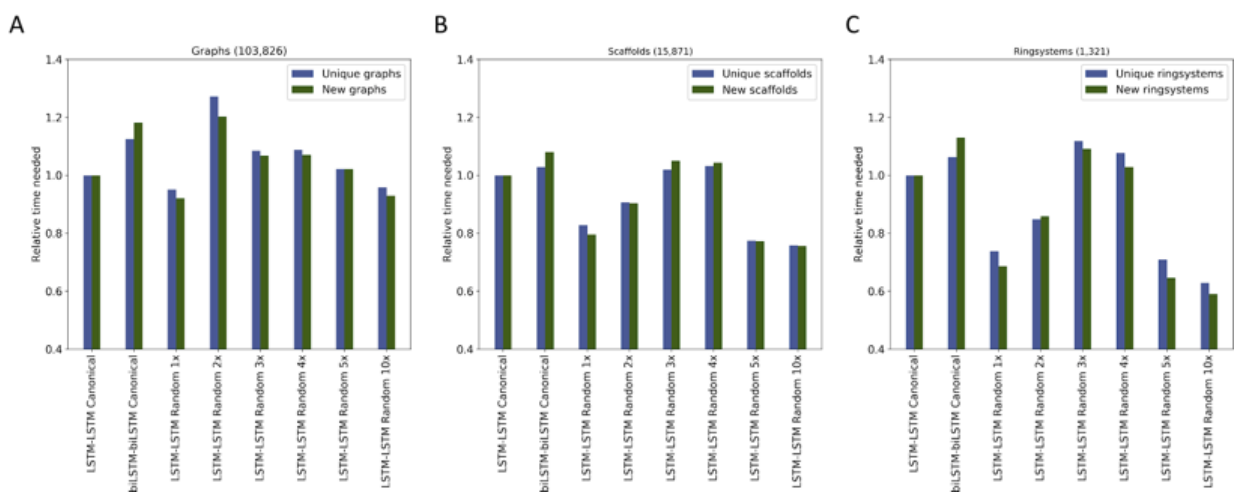

**Figure 6: Relative time needed to identify new graphs, scaffolds and rings.** The relative time is measured against the architecture using LSTM-LSTM with canonical graphs. A) Results for graphs. B) Results for scaffolds. C) Results for ring systems.

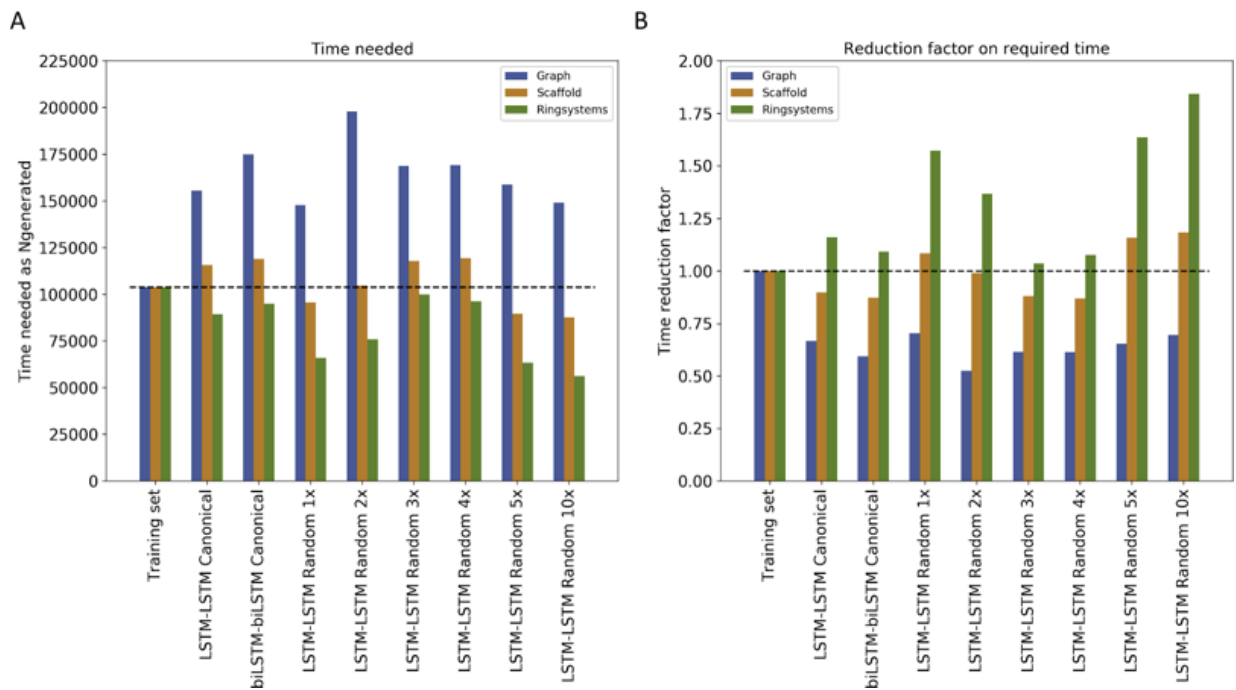

**Figure 7: Comparison of generated datasets with training set to identify new graphs, rings and scaffolds.** The dashed line indicates the training set. A) Time needed to reach the same number of unique graphs (blue), scaffolds (orange) and ring systems (green) as observed in the training set. B) Reduction factor on the time needed for graphs (blue), scaffolds (orange) and ring systems (green).

**Discussion**

Herein, we propose new pipeline to generate graphs with very high novelty, discovery and validity. This pipeline is inspired by SMILES generation models using GEN [27]. To our knowledge, the approach presented herein is the first text-based approach to generative sparse graphs using a deep generative model that can learn from existing sets of graphs [1]. An additionally novelty to sparse graph modelling has been introduced with the introduction of the LGI-format. This graph representation format has the additional benefit that it can randomized and augmented using multiple randomization on the same input graph. The possibility the augment the graphs may be beneficial for graph modelling on smaller graph datasets. Thirdly, the LGI-format provides a short text-based format for graph representation and may render themselves useful for compressed storage of undirected unweighted graphs. An additional benefit of the short format includes the possibility of modelling large graph datasets. Herein we have modelled a dataset of only 103k different graphs. During augmentation the format was increased to >1M LGI-strings which were subsequently modelling using our GEN-approach. This shows that the format and approach are open to very large datasets. Summarizing, introducing the one-line text format LGI allowed us to find an adequate answer to the challenges mentioned in the introduction and can address both size.

Our interest was to prove that we can explore sparse graphs complexity without exhaustively enumerating graphs with a tool such as GENG [22]. In particular, augmentation is essential to reproduce dataset knowledge and with a high degree of uniqueness and novelty. We have tested an online augmentation process which generate new random SMILES per molecule at each epoch, this feature is integrated in OCHEM for textCNN and textCNF methods. There are two main drawbacks of the online augmentation. First, before each epoch you need to generate a pool of new SMILES. Second, this method did not reduce the number of epochs to get an optimal solution. To speed up the first issue, a semi on-line augmentation was proposed which load a stored pregenerated list of SMILES of each molecule per epochs (JLR). Due to the very fast convergence of our models we cannot use those offline or semi offline methods which need lot of epochs to converge. Also, for prediction of properties, we observed similar results between online and off-line augmentations, so these two other methods were not tested.

In this study, we have shown that offline augmentation is also beneficial to LGI strings compared to SMILES strings. This was observed in several methods on SMILES strings, for example in prediction of reactions, for prediction of properties and for prediction of retrosynthesis path [44,45]. As our results suggest, the number of augmentations strongly influences the model performance. We therefore encourage the use of at least 3-fold augmentation to improve the models' performance. Further research is needed to understand the effect of augmentation on neural networks.

In the present study, we used GEN with examination of the validity of the LGI strings produced by the generator, this can be extended or combined to more general functions, like graph energy similarity and/or number of nodes ||V|| similarity for example. Another very interesting technique not tested here is the transfer learning. We have train our models in a given dataset but we can freeze part of our network and reuse this model to generate graph from another dataset using GEN.

We currently use the examinator to adjust early stopping procedure to stop models if our target is achieved. The Callback function is open to other modifications including of change of the learning rate and learning rate scheduling. Learning can be easily continued, suggesting the GEN-framework is also open to transfer learning if a different input dataset is chosen [46].

Our proposed LGI string was strongly inspired by SMILES generator tools to avoid reengineering graph to string conversion. This shows that the LGI- and SMILES-formats can be rapidly adapted to sparse graph problems in other fields of science. As displayed by the LGI-format and SMILES-format for graphs and molecules respectively, this type of one-line text representation for graphs is highly versatile and can describe different types of vertices. The format has excellent performance in text-based machine learning using neural networks for generative networks. We expect that these formats are also good representations for classification and regression networks as has been previously demonstrated for molecule using the SMILES representation. For sparse graphs used in other fields of science the vertex based LGI-format is open to machine learning assigning different vertex types to different ASCII characters. As shown with SMILES format for molecule, the format can also be adapted to include weighted edges.

**Conclusion**

To our knowledge, this is the first time that graphs are represented in strings (g6 or LGI) and used to learn generative models for sparse graphs. Additionally, the chosen LGI-format is augmentable and it is the first time an augmentable format for graph presentation has been used. Furthermore, it has been shown that augmentation is essential for improvement of the generative models. Finally, this is also the first time our recently introduce Generative Examination Networkx (GEN) are used to create a generative model for graphs. The combination of those three novel aspects allow us to make very fast models which can be trained with small and very large sets of known graphs. The model generates graph with a high percentage of validity, uniqueness and novelty allowing us to discover new scaffolds and ring systems. We have two opposite phenomena during the generation of data, i.e. reproduce the knowledge domain vs. novelty.

We have shown that both g6 and LGI strings can be used for graph representation but we know that one may propose another text format. So graph can be translated to any type of format and the characters may present different features, but the results that augmentation is an essential feature to improve the results for text models on graph. Non-augmentable formats such as InchiKeys for molecules perform poorly. We also know that g6 augmentation do not maintain enough local topology information which is essential for successful augmentation models.

## Declarations

### Availability of data and materials

The code and datasets used for this research are available on GitHub:

https://github.com/RuudFirsa/Graph-GEN

### Funding

RvD and GG are full-time employees of Firmenich SA, Geneva, Switzerland and funded by internal sources.

### Acknowledgement

The authors thank Prof.Dr. Pierre Vandergheynst (École Polytechnique Fédérale de Lausanne, Switzerland) and Sven Jeanrenaud (Firmenich SA) for critically review of the manuscript and suggested improvements.

### Competing interests
The authors declare that they do not have any competing interests.

### Contributions

RvD introduced the LGI-format and has written all code for the architecture, learning and generation. RvD also performed all calculations and analysis using both g6- and LGI-format. RvD and GG contributed equally to all experiments and publication.

**Supporting Information:**

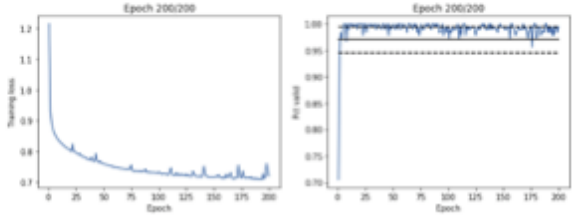

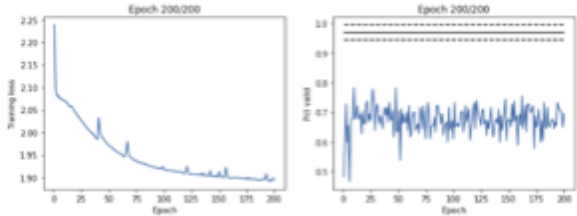

**Figure SI1: Results g6-strings during training.** The samples are based on a statistical sample of 180 generated graphs after every epoch. A) Results on generated graphs using a canonical g6-string; B) Results on generated graphs using a random g6-string.

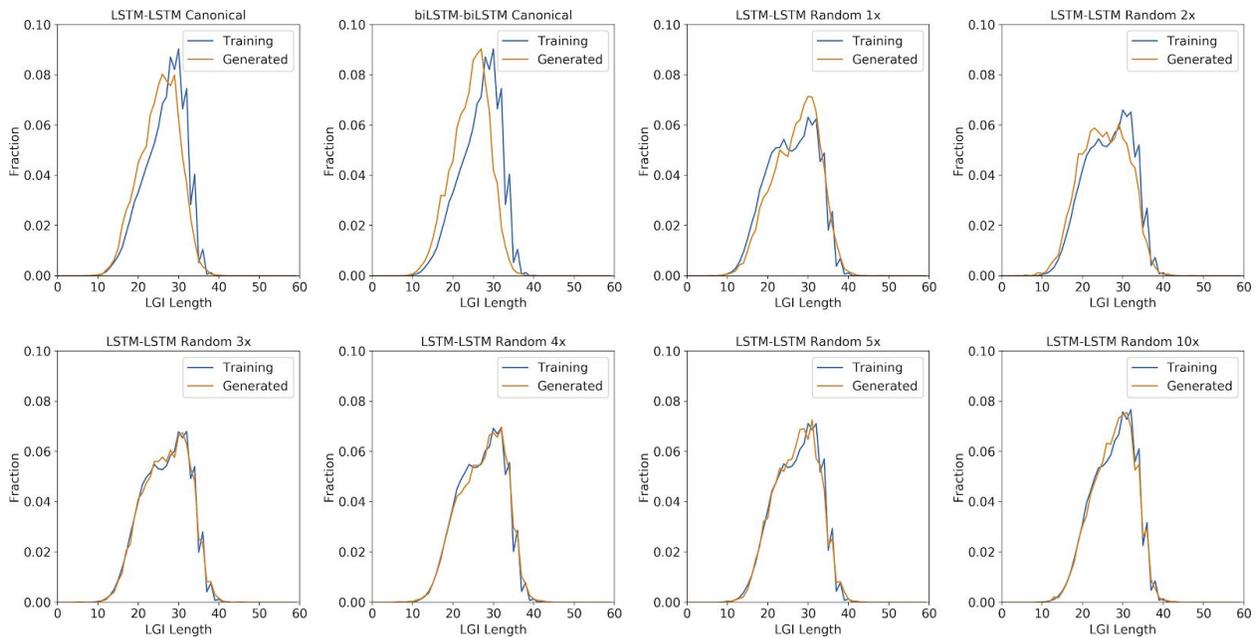

**Figure SI2: Distributions for LGI-length.** The data shows the results for LSTM-LSTM canonical, biLSTM-biLSTM canonical and the augmented random datasets with augmentations of 1, 2, 3, 4, 5 and 10 random LGI-strings per input graph.

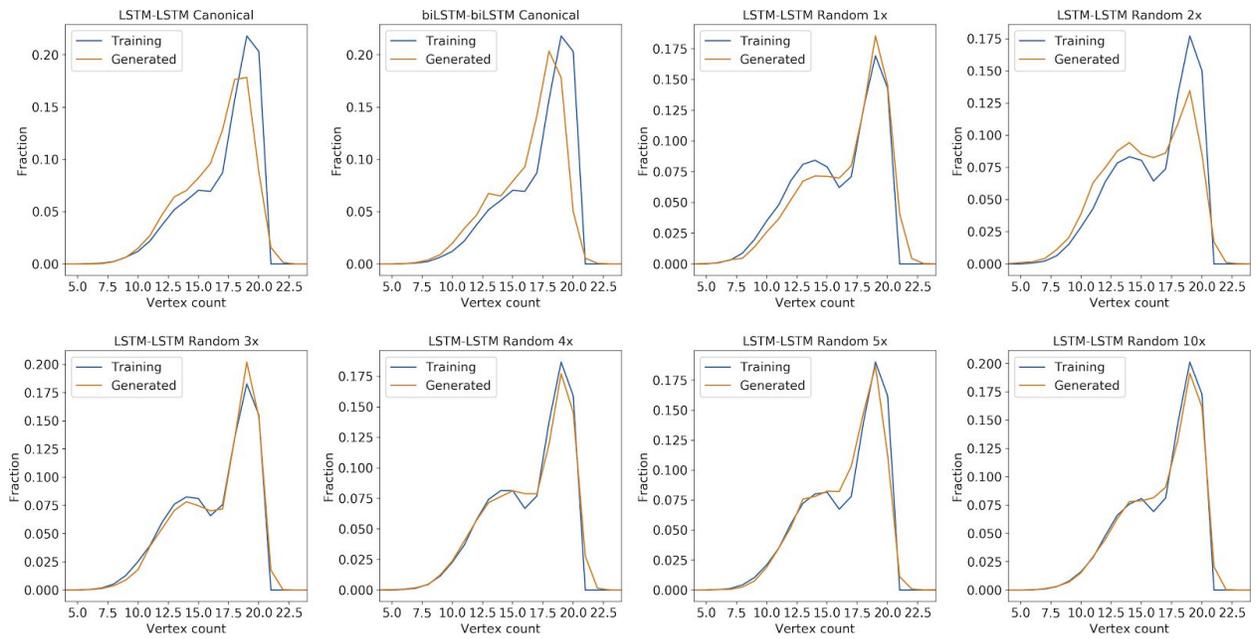

**Figure SI3: Distributions for vertex count in the training and generated spaces.** The data shows the results for LSTM-LSTM canonical, biLSTM-biLSTM canonical and the augmented random datasets with augmentations of 1, 2, 3, 4, 5 and 10 random LGI-strings per input graph.

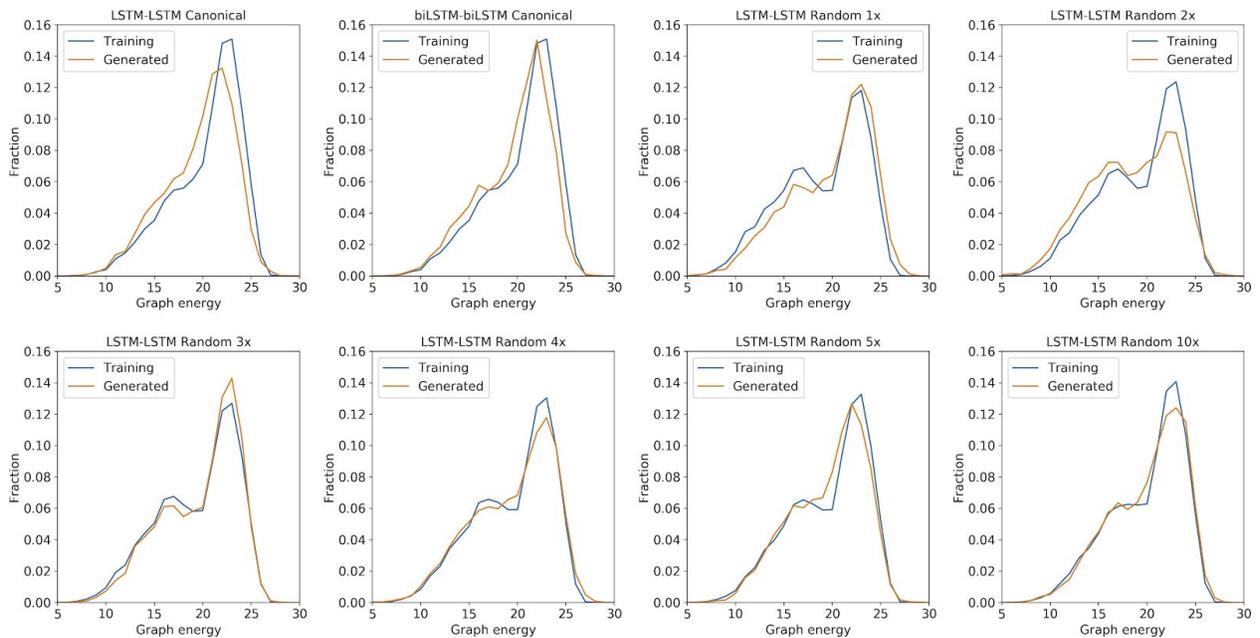

**Figure SI4: Distributions for graph energies GE in the training and generated spaces.** The data shows the results for LSTM-LSTM canonical, biLSTM-biLSTM canonical and the augmented random datasets with augmentations of 1, 2, 3, 4, 5 and 10 random LGI-strings per input graph.

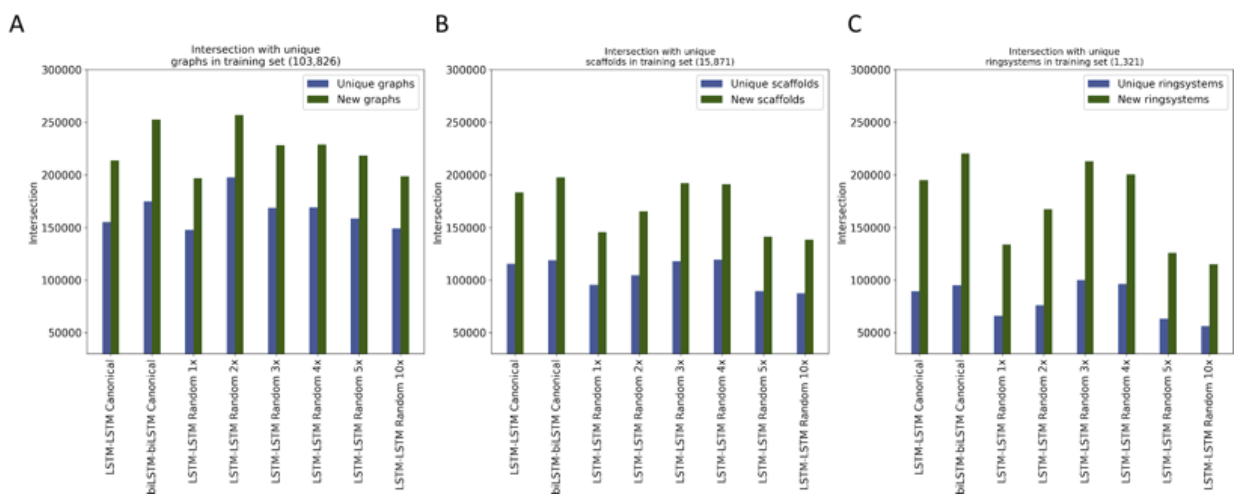

**Figure SI5: Intersection for graphs, scaffolds and ring systems with the training set.** The computed intersection indicate the number of compounds where the set of generated unique graphs was equal to the number of unique training compounds. "Unique" refers to unique generated graphs. "New" refers to unique generated graphs unknown in the training dataset. A) Results for graphs. B) Results for scaffolds. C) Results for ring systems.